\documentclass[letterpaper,10pt,conference]{ieeeconf}  
\usepackage{times}

\IEEEoverridecommandlockouts                              
\usepackage[utf8]{inputenc}
\usepackage[T1]{fontenc}
\makeatletter %
\let\NAT@parse\undefined %
\makeatother

\usepackage{amsmath} 
\usepackage{amssymb}  
\usepackage{tikz}
\usepackage{mathrsfs} 
\usepackage{pifont}   
\usepackage{amsfonts}

\usepackage[caption=false,font=footnotesize]{subfig}
\usepackage{graphicx}
\usepackage{xcolor}
\usepackage{multirow}
\usepackage{wasysym}
\graphicspath{{fig/}}
\usepackage[binary-units]{siunitx}[=v2]
\usepackage{booktabs}

\newcommand{\algrule}[1][.4pt]{\par\vskip.2\baselineskip\hrule height #1\par\vskip.2\baselineskip}
\usepackage[ruled,vlined,linesnumbered]{algorithm2e} 
\SetKwInOut{Parameters}{Parameters}
\usepackage[
   linkcolor={red!50!black},
   citecolor={blue!50!black},
   urlcolor={blue!80!black},
   colorlinks,
   bookmarks=false,
   pdfusetitle,
]{hyperref}

\usepackage[capitalise]{cleveref}
\Crefformat{figure}{#2Fig.~#1#3}
\Crefformat{equation}{(#2#1#3)}
\Crefmultiformat{figure}{Figs.~#2#1#3}{ and~#2#1#3}{, #2#1#3}{ and~#2#1#3}
\Crefrangeformat{equation}{(#3#1#4--#5#2#6)}
\Crefmultiformat{equation}{(#2#1#3)}{ and~(#2#1#3)}{, (#2#1#3)}{ and~(#2#1#3)}

\newcommand{\disable}[1]{}

\newcommand{\doi}[1]{\textsc{doi}: \href{https://doi.org/#1}{\nolinkurl{#1}}}

\begin{document}

\title{\LARGE \bf
LiLi: Lie Theory Based 3D LiDAR \\
Scan Alignment Degeneracy Detection
}

\disable{
   \urldef{\mails}\path|{,faiglj}@fel.cvut.cz|    
   {\tt\small \{hulchvse|bayerja1|faiglj\}@fel.cvut.cz}
   \authors{
      and Vsevolod Hulchuk\orcidID{0009-0003-8809-3052}
      \and Jan Bayer\orcidID{0000-0003-1190-1085}
      \and Jan Faigl\orcidID{0000-0002-6193-0792}}
   \authorrunning{V. Hulchuk et al.}
   \institute{Faculty of Electrical Engineering, Czech Technical University in Prague,\\
   Technická, 2, 166 27, Prague, Czechia\\
   \url{https://comrob.fel.cvut.cz}}
}

\author{Vsevolod Hulchuk\texorpdfstring{\authorrefmark{1}}{} \and Jan Bayer\texorpdfstring{\authorrefmark{1}}{} \and Jan Faigl
\thanks{\authorrefmark{1}These authors contributed equally to this work.}%
\thanks{The authors are with the
Faculty of Electrical Engineering,
Czech Technical University,
Technická 2, 166 27, Prague,
Czech Republic
   {\tt\small \{hulchvse|bayerja1|faiglj\}@fel.cvut.cz}
}%
}
\maketitle

\begin{abstract}
In this paper, we study 3D LiDAR scan alignment in challenging scenarios with degeneracies, such as straight corridors or flat fields, where the alignment solution is not unique and compromises localization and mapping accuracy.
Existing degeneracy detection methods that neglect the potential for reassociating data points are prone to being sensitive to noise and complex degeneracies.
Therefore, we propose LiLi -- a novel method that leverages Lie theory to identify the full set of degenerate transformations within the $SE(3)$ Lie group of rigid transformations.
The method employs perturbations of the optimized solution and compares the resulting optimized poses to ensure robust detection of degeneracies.
By leveraging generators from the Lie algebra $\mathfrak{se}(3)$, the method provides a systematic approach to describing the set of degenerate transformations.
Quantitative evaluations on synthetic data show significant improvement over the state-of-the-art Hessian-based method, reducing alignment error by \SI{50}{\percent}, with more significant improvements for datasets featuring noise.
In the real-world degenerate datasets, the proposed method integrated into LiDAR-based odometry yields superior localization performance compared to the reference solution based on the Hessian-based degeneracy detector on a \SI{260}{\meter} long trajectory, and succeeds on a \SI{430}{\meter} long round-trip tunnel trajectory where the reference fails.

\end{abstract}
\section{Introduction} \label{sec:introduction}
\textit{3D LiDARs} are widely used sensors that provide high-precision 3D point cloud scans of the environment.
The alignment of consecutive scans is essential for autonomous navigation when 3D LiDAR scan alignment is used to determine the relative transformation between scans or a~scan and a map.
It computes the 6D transformation (3D translation and 3D rotation) between scans.
Hence, it is a foundational component of LiDAR-based localization and mapping.
However, achieving robust alignment becomes challenging in environments where geometric constraints are insufficient, leading to degeneracies.

\begin{figure}[!htb]
   \centering
   \subfloat[Synthetic data results. Local scan (blue), aligned against the reference scan (red). Yellow: LiLi-based scan extension~\cref{eq:extended_scan} (proposed, Perturbation-based), green: Zhang's-based scan extension (state-of-the-art, Hessian-based) \label{fig:teaser_synthetic}]{%
      \includegraphics[width=0.48\textwidth]{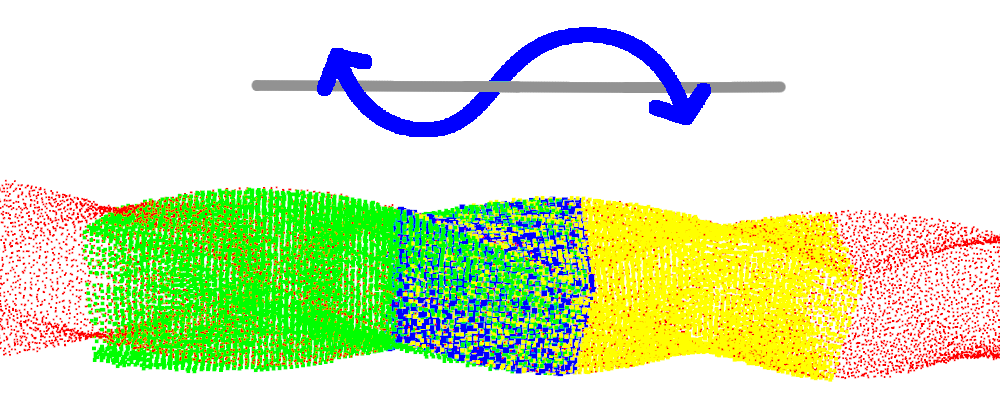}%
   }
   \hfill 
   \subfloat[Real-world degenerate scan matching of local scan (blue) against the accumulated map (red) from the custom Tunnel dataset. Localization is in progress, supported by the proposed LiLi degeneracy detection, resulting in the gray trajectory. 3D axes represent the currently estimated robot pose\label{fig:teaser_real}]{%
      \includegraphics[width=0.48\textwidth]{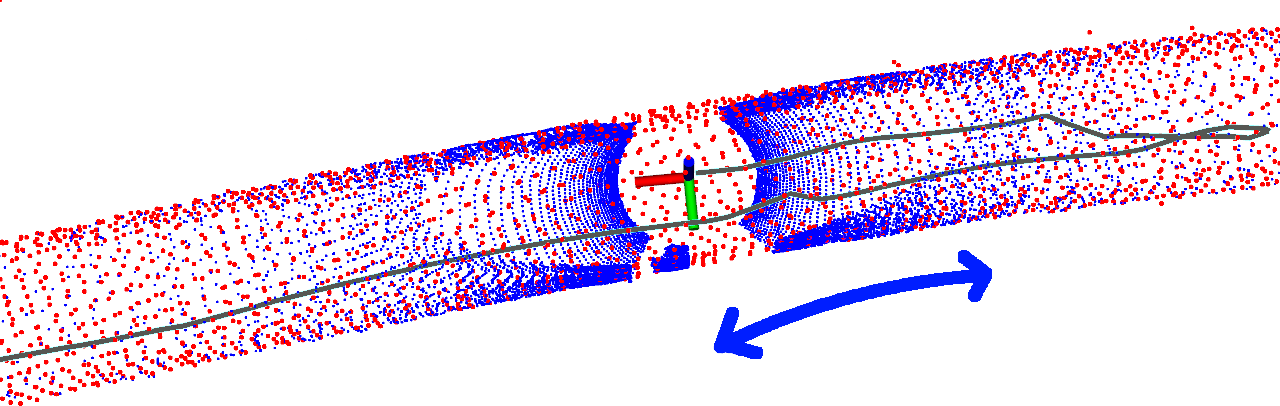}%
   }
   \caption{
      Examples of complex, coupled degeneracy in 3D LiDAR scan alignment on \protect\subref{fig:teaser_synthetic} synthetic and \protect\subref{fig:teaser_real} real-world tunnel data. 
      In both cases, a structurally degenerate direction (blue arrow), composed of translational and rotational components, preserves the alignment between the reference scan (red) and the target scan (blue).
      On the synthetic data, the local scan extension outperforms the baseline. On real-world data, the proposed method enables robust robot localization, while the baseline determines the degeneracy inconsistently, leading to localization failure.
   }
   \label{fig:teaser}
\end{figure}

Degeneracy refers to scenarios where specific movements, such as translations, rotations, or combinations thereof, do not significantly affect the perceived alignment due to a lack of distinctive structural features in the scene.
For instance, a~scan taken in the middle of an open field can be translated along the ground plane or rotated about the vertical axis without significantly affecting the measure of the alignment quality.
More complex cases involve coupled translational and rotational motions, as depicted in \cref{fig:teaser}, which showcases such degeneracy on both synthetic and real-world data.

We consider understanding the space of degenerate motions important because it directly affects the reliability of alignment algorithms. 
These motions represent degrees of freedom that are not constrained by the scene's geometry, leading to ambiguities in pose estimation.
Identifying the specific types of degeneracies, such as planar translations or axis-aligned rotations, helps in designing methods to mitigate their effects.
Addressing these issues improves the robustness of alignment systems, especially in feature-poor environments.

Existing methods that focus on analyzing the final alignment ignore the potential of reassociating the data points.
While effective under ideal conditions, such methods may fail to account for degenerate motions that include re-association of data points, especially in noisy and cluttered environments. 
To address this limitation, we introduce LiLi (Lie Theory Based 3D LiDAR Scan Alignment Degeneracy Detection). Our method leverages Lie theory to characterize the degenerate subspace of rigid motion transformations by analyzing the alignment system's response to controlled perturbations, explicitly handling the re-association of points to ensure robustness.

The contributions of this work are considered as follows.
\begin{itemize}
    \item A novel formulation of scan alignment degeneracy grounded in Lie theory, which describes the degenerate subspace using generators from the Lie algebra, $\mathfrak{se}(3)$.
    \item The LiLi algorithm, a perturbation-based method that robustly detects and describes the degeneracy by explicitly accounting for the re-association of data points.
    \item The method validation on synthetic and real-world datasets, demonstrating a significant improvement over the state-of-the-art, especially with the presence of noise, and enabling robust localization in scenarios where the state-of-the-art method fails completely.
\end{itemize}

The remainder of the paper is organized as follows.
\cref{sec:related} provides an overview of related work on degeneracy detection and scan alignment.
The proposed scan alignment degeneracy description using Lie theory is presented in \cref{sec:deg}.
Upon Lie theory formulation of the degeneracy in scan alignment, the problem formulation is presented in \cref{sec:problem}.
The proposed method is then detailed in \cref{sec:method}.
Evaluation methodology and experimental results are described in \cref{sec:results}.
A discussion of the found insights is covered in \cref{sec:discussion}.
Finally, Section~\ref{sec:conclusion} concludes the paper with a summary of our findings and potential future directions.

\section{Related Work} \label{sec:related}

The LiDAR scan alignment degeneracy is a common challenge in scan alignment algorithms, particularly in point-to-plane methods, which are widely used in LiDAR-based SLAM systems, such as LOAM~\cite{zhang2014loam}.
The general principle of scan alignment is to find the optimal rotation $\mathbf{R}$ and translation $\mathbf{t}$ by minimizing the following cost function.
\begin{equation}\label{eq:argmin_residuals}
    \begin{aligned}
        \mathcal{C}(\mathbf{R}, \mathbf{t}) = \sum_{i} \| \mathbf{r}_i \|^2,
    \end{aligned}
\end{equation}
where $\mathcal{C}(\mathbf{R}, \mathbf{t})$ is the cost function, and $\mathbf{r}_i$ represents the residual for the $i$-th correspondence.
Specifically, in the point-to-plane alignment, the residual is $\mathbf{r}_i = \mathbf{n}_i^\top (\mathbf{R} \mathbf{p}_i + \mathbf{t} - \mathbf{q}_i)$, where $\mathbf{n}_i$ is the normal vector at the corresponding plane, $\mathbf{p}_i$ is the point in the source point cloud, and $\mathbf{q}_i$ is the point in the target point cloud.

One foundational approach in degeneracy detection of the LiDAR scan alignment is \textit{Zhang's degeneracy detection}~\cite{zhang2016degeneracy}.
It formulates degeneracy as a part of the general optimization-based problem, which is solved by local linearization.
Nonlinear solvers linearize the cost function~\cref{eq:argmin_residuals} by computing the Jacobian of the residual vector $\mathbf{r}$ with respect to the optimized transformation parameters ($\mathbf{R}, \mathbf{t}$),
\begin{equation}\label{eq:jacobian}
    \begin{aligned}
        \mathbf{J} = \frac{\partial \mathbf{r}}{\partial [\mathbf{R}, \mathbf{t}]},
    \end{aligned}
\end{equation}
which yields the linearized system $\mathbf{A} \mathbf{x} = \mathbf{b}$, where $\mathbf{A}$ is the weighted Jacobian retaining the weights of the original problem.
The authors of~\cite{zhang2016degeneracy} analyze the eigenvalues of $\mathbf{A}^\top \mathbf{A}$, and identify the directions in which the smallest eigenvalues fall below a threshold as degenerate, leaving the solution in those directions unchanged by \emph{solution remapping}.
For the least-squares cost~\cref{eq:argmin_residuals}, the analyzed matrix
\begin{equation}\label{eq:hessian}
    \begin{aligned}
        \mathbf{A}^\top \mathbf{A} \approx \mathbf{H}
    \end{aligned}
\end{equation}
is the Gauss--Newton approximation of the Hessian $\mathbf{H}$ of the cost function, and we thus refer to such methods as \emph{Hessian-based}.
While computationally efficient, \textit{Zhang's degeneracy detection} assumes static correspondences and does not explicitly account for dynamic data reassociation during the optimization process. 
Since this limitation makes Zhang's method sensitive to noise and less robust in complex or noisy environments where point cloud associations evolve iteratively, \emph{we propose to address the limitation in the developed method}.

Further research has sought to enhance robustness by repeating the eigenvalue analysis at each step of the ICP algorithm~\cite{hinduja2019degeneracy} or incorporating degeneracy-aware factors for graph optimization~\cite{ren2020towards}.
Other efforts have aimed to minimize degeneracy by selecting features that reduce the degeneracy measure~\cite{wang2022towards}, applying the method to single-scan observability analysis~\cite{denniston2022loop}, or evaluating degeneracy using the log-determinant of the Information matrix spectrum~\cite{jiao2021greedy}. 
While these methods improve the robustness of the ICP and can be incorporated in global alignment, \emph{their handling of noise and data reassociation remains limited}.

Degeneracy detection has also been extensively studied within the ICP framework through information matrix analysis.
For instance, \cite{censi2007achievable} proposed analyzing the information matrix to detect degeneracies, building on earlier work~\cite{censi2007accurate} that describes covariance estimation for 2D LiDAR scans.
More recently, \cite{talbot2023principled} improved upon these methods by incorporating measurement uncertainty and applying the approach to geometrically degenerate environments, such as lunar surfaces.
However, \emph{these methods do not directly address the data reassociation challenges for degeneracy detection in LiDAR scan alignment}.

Another perspective focuses on the distribution of planar feature normals.
For example, \cite{nashed2021robust} introduced a soft-constraint regularization term to prevent optimization along degenerate directions based on the normal distribution.
Similarly, \cite{zhen2017robust} and \cite{nobili2018predicting} used the eigenvalue proportions of the covariance matrix of normals to predict degeneracy and scan ``alignability,'' with \cite{ramezani2020online} proposing an online adaptation of the method for a robust SLAM.
These \emph{methods are sensitive to noise as it directly affects the estimated directions of the normals}. 
Besides, methods that use the covariance matrix of 3D normals, \emph{are limited to extract only translational degeneracies}. 

Alternative strategies have also emerged to detect or tackle degeneracies. 
For instance, \cite{tuna2023x} analyzed the torques induced by constraints around a virtual wrench to detect degeneracy, while neural network-based approaches~\cite{nubertlearning,wang2022lidar} show promise in efficiently identifying degeneracies.
Some methods, such as those described in \cite{nakamura2021short} and \cite{hulchuk2023graph}, detect degeneracy by comparing LiDAR positions with predictions from external sources, such as odometry or IMU.
In \cite{baril2022kilometer}, the authors explicitly analyze the sensitivity of ICP to the pose perturbations. However, the method \emph{is not designed to describe the degeneracy space}, but to measure the overall standard deviation of the pose.
Other approaches bypass degeneracy detection altogether by complementing LiDAR scans with additional data, as seen in R3Live~\cite{lin2022r}, which combines LiDAR scans with visual and inertial data for robust localization, or HECTOR SLAM~\cite{nagla20202d}, which utilizes motion models to address corridor degeneracies for 2D LiDAR scans alignment.

While significant progress has been made in detecting and mitigating degeneracies, key challenges remain. 
Notably, handling dynamic data reassociation and noise in coupled rotational and translational degeneracies continues to pose difficulties.
Our work seeks to address these gaps by providing a systematic framework for degeneracy detection and representation, advancing the state-of-the-art in LiDAR scan alignment.

\section{Degeneracy in 3D LiDAR Scan Alignment}\label{sec:deg}

In this section, we formulate the scan alignment degeneracy based on Lie theory to explicitly characterize the continuous space of degenerate transformations in scan alignment.
3D LiDAR scan alignment estimates the transformation between two scans or between a scan and a map.
Degeneracy arises when there exists a continuous set of ways to align the scans with negligible differences in the alignment quality. 

Formally, degeneracy is defined in~\cite{zhang2016degeneracy} as a property of the optimization problem underlying scan alignment, expressed as
\begin{equation}\label{eq:argmin}
   \begin{aligned}
      \underset{\mathbf{x}}{\text{argmin}} \quad f^2(\mathbf{x}),
   \end{aligned}
\end{equation}
where $f(\mathbf{x})$ is the cost function that measures the alignment error.

The degeneracy factor $\mathcal{D}$ along a direction $\mathbf{c}$ in the \num{6}~degrees of freedom parameter space quantifies the sensitivity of the solution to the perturbations $\delta \mathbf{d}$ as
\begin{equation}\label{eq:degeneracy}
   \begin{aligned}
      \mathcal{D} = \frac{\delta \mathbf{d}}{\delta \mathbf{x}_c},
   \end{aligned}
\end{equation}
where $\delta \mathbf{d}$ represents the amount of perturbation applied, and $\delta \mathbf{x}_c$ measures the corresponding change in the solution.

\subsection*{Degenerate Directions in $\mathfrak{se}(3)$ and Transformations in $SE(3)$}

Building on the definition \cref{eq:degeneracy}, the space of degenerate local twists can be defined using the Lie algebra of twists $\mathfrak{se}(3)$, which represents infinitesimal motions in 3D space, capturing both translational and rotational components.
Twists in $\mathfrak{se}(3)$ represent degenerate directions in the parameter space that generate paths of degenerate transformations in the Lie group $SE(3)$.

A twist $\mathbf{x}_{\mathrm{degeneracy}}$ can be exponentiated using the Lie theory exponential map to generate a continuous path of transformations in $SE(3)$ as
\begin{equation}
   \mathbf{T}(t) = \exp(t \cdot \mathbf{x}_{\mathrm{degeneracy}}),
\end{equation}
where $t$ is a scalar parameter. 
It produces the trajectory of motions described by the twist.

We can identify a set of basis twists $\mathbf{b}_1, \mathbf{b}_2, \ldots, \mathbf{b}_k$ that span the degenerate subspace in $\mathfrak{se}(3)$ to capture all possible degenerate motions.
The complete set of degenerate transformations in $SE(3)$ is then given by
\begin{equation}\label{eq:degeneracy_span}
   \mathcal{T}_{\mathrm{degenerate}} = \{ \exp\left(\sum_{j=1}^{k} c_j \cdot \mathbf{b}_j\right) \mid c_j \in \mathbb{R} \},
\end{equation}
where $\mathcal{T}_{\mathrm{degenerate}}$ is the full set of degenerate transformations in $SE(3)$, $\{\mathbf{b}_j\}_{j=1}^{k}$ are the basis twists spanning the degenerate subspace in $\mathfrak{se}(3)$, and $c_j$ are the coefficients defining the linear combinations of the basis twists.
It ensures that all degenerate transformations consistent with the local geometry are captured.

\section{Problem Statement} \label{sec:problem}
Given a reference scan, a local scan, the initial position of the local scan, and the alignment algorithm, the goal is to determine a set of basis twists in $\mathfrak{se}(3)$ that span the subspace of degenerate transformations in $SE(3)$.
Through applying an exponential map to the linear combinations of the basis twists~\cref{eq:degeneracy_span}, all the degenerate transformations observable within the local geometry of the scans are generated.

We propose the following method to evaluate the quality of the determined degenerate directions.

\subsection*{Proposed Quality Measurement}

We extend the local scan of the surface by applying combinations of the detected basis twists over time and stacking the corresponding point clouds.
The process generates an extended representation of the local scan.
The extended scan is computed as
\begin{equation}\label{eq:extended_scan}
   \mathbf{S}_{\text{extended}} = \bigcup_{i=1}^{n} \mathbf{S}_{\text{local}} \cdot \mathbf{T}_i,
\end{equation}
where $\mathbf{S}_{\text{local}}$ is the original local scan, $n$ is the number of samples of degenerate motions, and $\mathbf{T}_i$ represents the transformations generated by combining the optimized initial position $\mathbf{P}_0$ and the span of the detected basis twists
\begin{equation}
   \mathbf{T}_i = \mathbf{P}_0 \cdot \exp\left( \sum_{j=1}^{k} c_{ij} \cdot \mathbf{b}_j \right),
\end{equation}
with $c_{ij}$ being the combination coefficients and $k$ the number of basis twists in the degenerate subspace.

The alignment quality is measured by comparing the extended local scan $\mathbf{S}_{\text{extended}}$ with the reference scan $\mathbf{S}_{\text{reference}}$.
Specifically, we evaluate the median distance from points on $\mathbf{S}_{\text{extended}}$ to their nearest neighbors in $\mathbf{S}_{\text{reference}}$ as
\begin{equation}\label{eq:quality}
   Q = \underset{\mathbf{p}_i \in \mathbf{S}_{\text{extended}}}{\text{median}}\big(\text{dist}(\mathbf{p}_i, \mathbf{S}_{\text{reference}})\big),
\end{equation}
where $\text{dist}(\mathbf{p}_i, \mathbf{S}_{\text{reference}})$ is the distance from a point $\mathbf{p}_i$ to its nearest neighbor in $\mathbf{S}_{\text{reference}}$.

The proposed metric quantifies how well the detected basis twists capture the local structure of the reference scan.
In an ideal case, where the object exhibits perfect local geometries, $Q$ measures the alignment quality by indicating how accurately the basis twists span the degenerate subspace of the local geometry.

\section{Proposed Method} \label{sec:method}

The proposed method identifies degeneracies by applying a set of axis-aligned perturbations -- $(\Delta t_x, \Delta r_x)$, $(\Delta t_y, \Delta r_y)$, and $(\Delta t_z, \Delta r_z)$ -- to an optimized pose and analyzing the results of re-optimization.
The magnitudes for the translational and rotational components are selected adaptively to ensure these perturbations are significant enough to reveal degeneracies without being arbitrarily large.
The selection is governed by a single integer parameter, $k$, which defines a~target re-association distance.

For each perturbation, the magnitudes of translation and rotation are calculated independently to meet the target.
The magnitude is calibrated to displace a benchmark point (one located at the median distance from the scan center) such that its new nearest neighbor in the reference scan is $k$~points away from its original correspondence.
For instance, a value of $k=3$ means the perturbation is just big enough to make the benchmark point slide two neighbors away to get a new correspondence.
The proposed approach ensures the perturbation scale is automatically adjusted based on the local point cloud density.

While the initial perturbations are axis-aligned, the set is sufficient to detect arbitrary degenerate directions.
Even if a perturbation does not align with a degenerate direction, the re-optimization process naturally causes the pose to converge toward the nearest low-cost region, effectively ``sliding'' along the degeneracy and revealing its direction.
The resulting deviations from the original optimized pose are then analyzed to construct the basis of the degenerate subspace.


\setlength{\algomargin}{1.2em}
\let\oldnl\nl
\newcommand{\nonl}{\renewcommand{\nl}{\let\nl\oldnl}}
\newcommand\mycommfont[1]{\small \it\textcolor{black}{#1}} 
\SetCommentSty{mycommfont}
\SetKwComment{tcp}{$\rhd$~}{}

\begin{algorithm}[!htb]
   \caption{LiLi Degeneracy Detection Method}\label{alg:degeneracy_detection}
   \DontPrintSemicolon
   \KwIn{
      $\mathbf{P}_{\mathrm{init}}$ -- Initial pose estimate,
      $\mathcal{P} = \{(\Delta t_x, \Delta r_x)$, $(\Delta t_y, \Delta r_y)$, $(\Delta t_z, \Delta r_z)\}$ -- Set of perturbations.
   }
   \Parameters{
      Optimization function $f_{\mathrm{opt}}$,
      Threshold for median displacement $\tau_{\mathrm{displacement}}$,
      Threshold for PCA eigenvalues $\tau_{\mathrm{PCA}}$.
   }
   \KwOut{
      $\mathbf{S}_{\mathrm{degeneracy}}$ -- Sparsified degeneracy subspace basis.
   }
   \DontPrintSemicolon
   \algrule

   \nonl \tcp*[h]{Optimize pose to obtain $\mathbf{P}_{\mathrm{opt}}$}\;
   \nl$\mathbf{P}_{\mathrm{opt}} \gets f_{\mathrm{opt}}(\mathbf{P}_{\mathrm{init}})$\label{line:optimization}\;

   \ForEach{\rm perturbation $\Delta \mathbf{p}_i \in \mathcal{P}$}{
      \nonl \tcp*[l]{Apply perturbation and re-optimize} 
      \nl $\mathbf{P}_{\mathrm{perturbed}} \gets f_{\mathrm{opt}}(\mathbf{P}_{\mathrm{opt}} + \Delta \mathbf{p}_i)$\label{line:pertb_start}\;
      \nonl \tcp*[l]{Compute degeneracy transformation}
      \nl $\mathbf{T}_{\mathrm{degeneracy}} \gets \mathbf{P}_{\mathrm{opt}}^{-1} \cdot \mathbf{P}_{\mathrm{perturbed}}$\;
      \nonl \tcp*[l]{Extract degeneracy direction}
      \nl$\mathbf{t}_{\mathrm{degeneracy}} \gets \mathrm{log}(\mathbf{T}_{\mathrm{degeneracy}})$\;
      
      \If{\rm median displacement of points exceeds $\tau_{\mathrm{displacement}}$}{
	 Store $\mathbf{t}_{\mathrm{degeneracy}}$ in $\mathbf{D}_{\mathrm{degeneracy}}$\label{line:pertb_end}\;
      }
   }

      \nonl \tcp*[l]{Analyze degeneracy directions}
   \nl Perform PCA on $\mathbf{D}_{\mathrm{degeneracy}}$ to obtain eigenvectors and eigenvalues.\label{line:pca}\;
   Select eigenvectors with eigenvalues greater than $\tau_{\mathrm{PCA}}$ to form $\mathbf{S}_{\mathrm{PCA}}$.\label{line:pca_threshold}\;

      \nonl \tcp*[l]{Sparsify PCA subspace to improve interpretability}
   \nl $\mathbf{S}_{\mathrm{degeneracy}} \gets \mathrm{sparsify}(\mathbf{S}_{\mathrm{PCA}})$\label{line:sparsify}\;

   \Return $\mathbf{S}_{\mathrm{degeneracy}}$\;
\end{algorithm}
\vspace{-1.5em}
\footnotesize
\noindent\textsuperscript{*}The input perturbation set $\mathcal{P}$ is adaptively generated based on the parameter $k$, as described in \cref{sec:method}.
\normalsize

The overall degeneracy detection and description is outlined in \cref{alg:degeneracy_detection} and can be summarized as follows.
First, the scan alignment optimization is performed to obtain an optimized pose $\mathbf{P}_{\mathrm{opt}}$ (Line~\ref{line:optimization}, \cref{alg:degeneracy_detection}).
Perturbations are applied in specified combinations of translational and rotational directions.
For each perturbation, the system is re-optimized, and deviations are analyzed to detect degeneracy (Lines~\ref{line:pertb_start}--\ref{line:pertb_end}, \cref{alg:degeneracy_detection}).
Detected degeneracy directions are aggregated, and the \emph{Principal Component Analysis} (PCA) is applied to identify the degeneracy subspace (Line~\ref{line:pca}, \cref{alg:degeneracy_detection}), retaining only eigenvectors with eigenvalues above the threshold $\tau_{\mathrm{PCA}}$ (Line~\ref{line:pca_threshold}, \cref{alg:degeneracy_detection}).
Finally, sparsification is performed to improve interpretability (Line~\ref{line:sparsify}, \cref{alg:degeneracy_detection}).
The method is further detailed in the remaining part of the section.

\subsection*{Details of the Method}

The idea of the proposed method is based on the scan alignment system's sensitivity evaluation to a set of perturbations of the optimized solution to uncover potential degeneracies.
First, the optimization is performed to converge to a solution $\mathbf{P}_{\mathrm{opt}}$.
The perturbations $(\Delta t_x, \Delta r_x)$, $(\Delta t_y, \Delta r_y)$, $(\Delta t_z, \Delta r_z)$ are applied to the solution, and the optimization is performed again to obtain $\mathbf{P}_{\mathrm{perturbed}}$.

If the median displacement of the point in the re-optimized point cloud, normalized by the median perturbation displacement, exceeds the given threshold $\tau_{\mathrm{displacement}}$, the extracted direction is considered degenerate.
The degenerate transformation is then computed as
\begin{equation}\label{eq:degeneracy_transformation}
   \mathbf{T}_{\mathrm{degenerate}} = \mathbf{P}_{\mathrm{opt}}^{-1} \cdot \mathbf{P}_{\mathrm{perturbed}}.
\end{equation}

From the transformation $\mathbf{T}_{\mathrm{degenerate}}$, the degeneracy direction is extracted as an element of the Lie algebra $\mathfrak{se}(3)$ as
\begin{equation}\label{eq:degeneracy_direction}
   \mathbf{t}_{\mathrm{degeneracy}} = \mathrm{log}(\mathbf{T}_{\mathrm{degenerate}}).
\end{equation}

Detected degeneracy directions from all perturbations are aggregated, and the PCA is applied to perform the degeneracy subspace.
The PCA is applied to the covariance matrix of the detected directions
\begin{equation}\label{eq:pca_covariance}
   \mathbf{C} = \frac{1}{n} \sum_{i=1}^{n} (\mathbf{t}_i - \bar{\mathbf{t}})(\mathbf{t}_i - \bar{\mathbf{t}})^T,
\end{equation}
where $\mathbf{t}_i$ are the detected degeneracy directions, and $\bar{\mathbf{t}}$ is their mean.
Eigenvectors with eigenvalues above the threshold $\tau_{\mathrm{PCA}}$ form the basis of the degeneracy subspace.

The basis is sparsified by minimizing the $\ell_1$-norm of the basis matrix to enhance interpretability using
\begin{equation}\label{eq:sparsification}
   \underset{\mathbf{T}}{\text{minimize}} \quad \|\mathbf{TB}\|_1 \quad \text{subject to} \quad \mathbf{T}^T \mathbf{T} = \mathbf{I},
\end{equation}
where $\mathbf{B}$ is the matrix representing the original basis, and $\mathbf{T}$ is an orthogonal matrix that defines the transformation of the basis. The optimal basis transformation, $\mathbf{T}_{\text{opt}}$, yields the new sparsified basis represented by basis matrix $\mathbf{B}_{\text{sparse}} = \mathbf{T}_{\text{opt}}\mathbf{B}$.

It ensures that the sparsified basis retains interpretability while representing the same degeneracy subspace. For example, when applied to planar degeneracy, the sparsified basis ideally consists of pure translations along the plane and a rotation around the plane normal.

\section{Evaluation Results} \label{sec:results}

The proposed method for degeneracy detection has been evaluated on both synthetic and real experimental datasets.
The synthetic dataset consists of geometric shapes that represent diverse degeneracy scenarios with and without measurement noise.
We measure the quality of the detected degeneracy subspace using the local scan extension alignment quality metric~\cref{eq:quality}.
Furthermore, we demonstrate the method using a real dataset with two trials. We integrate the method into a LiDAR-Inertial localization system and evaluate its performance using standard localization metrics.

The localization system is based on the state-of-the-art LiDAR-Inertial localization algorithm~\cite{shan2020lio}, with wheel odometry used to constrain the pose along the detected degenerate directions.
In both synthetic and real-world scenarios, we compare the proposed Perturbation-based method \emph{LiLi} with the reference method, which is the state-of-the-art Hessian-based degeneracy detection, referred to as \emph{Zhang's}~\cite{zhang2016degeneracy}.

The results on synthetic data are presented in \cref{sec:results_synthetic}, and the performance of the method in the localization task is reported in \cref{sec:results_localization}.%
\footnote{The dataset would be made publicly available together with the reference implementation of the proposed method.
Due to the double-anonymous review for ICRA 2026, they are not disclosed here.} 

\subsection{Degeneracy Detection on Synthetic Data} \label{sec:results_synthetic}
\begin{figure}[h!]
    \centering
    \subfloat[Plane]{\includegraphics[height=2.5cm]{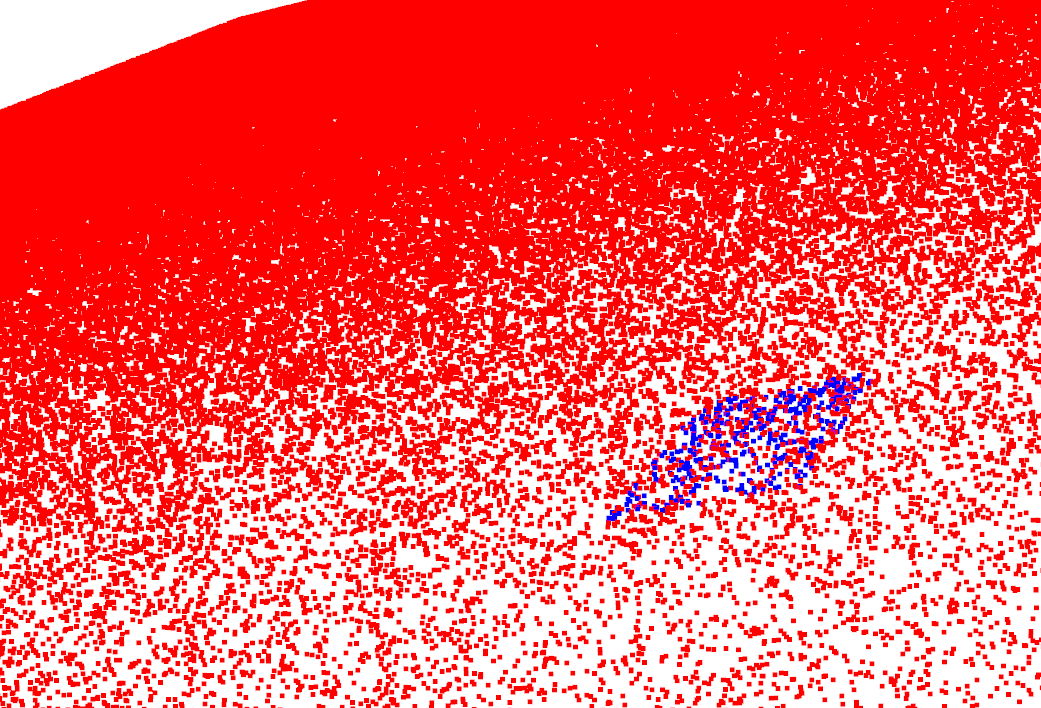}}
    \hfill 
    \subfloat[Closed Cylinder]{\includegraphics[height=2.5cm]{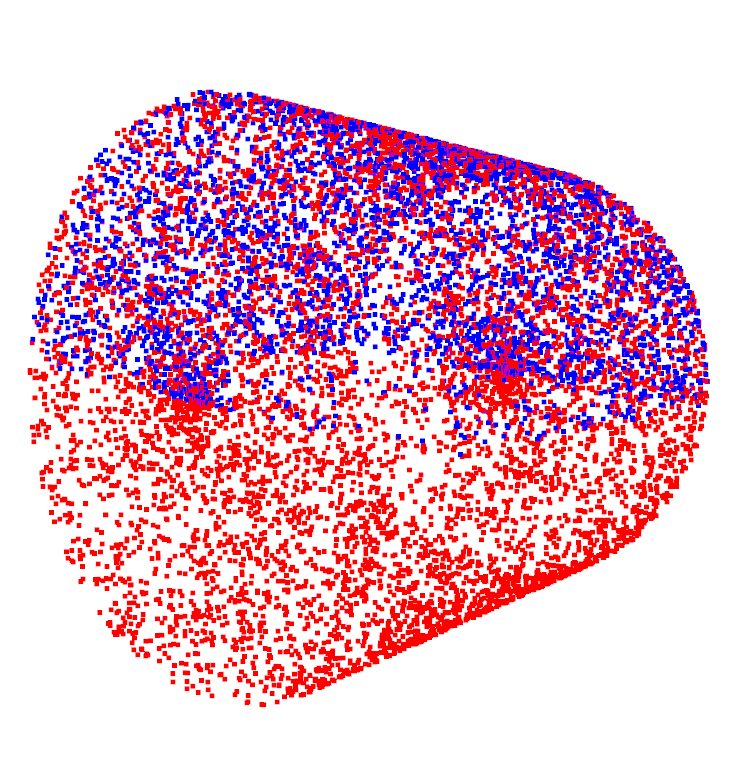}}

    \subfloat[Sinusoidal Cylinder]{\includegraphics[height=2.5cm]{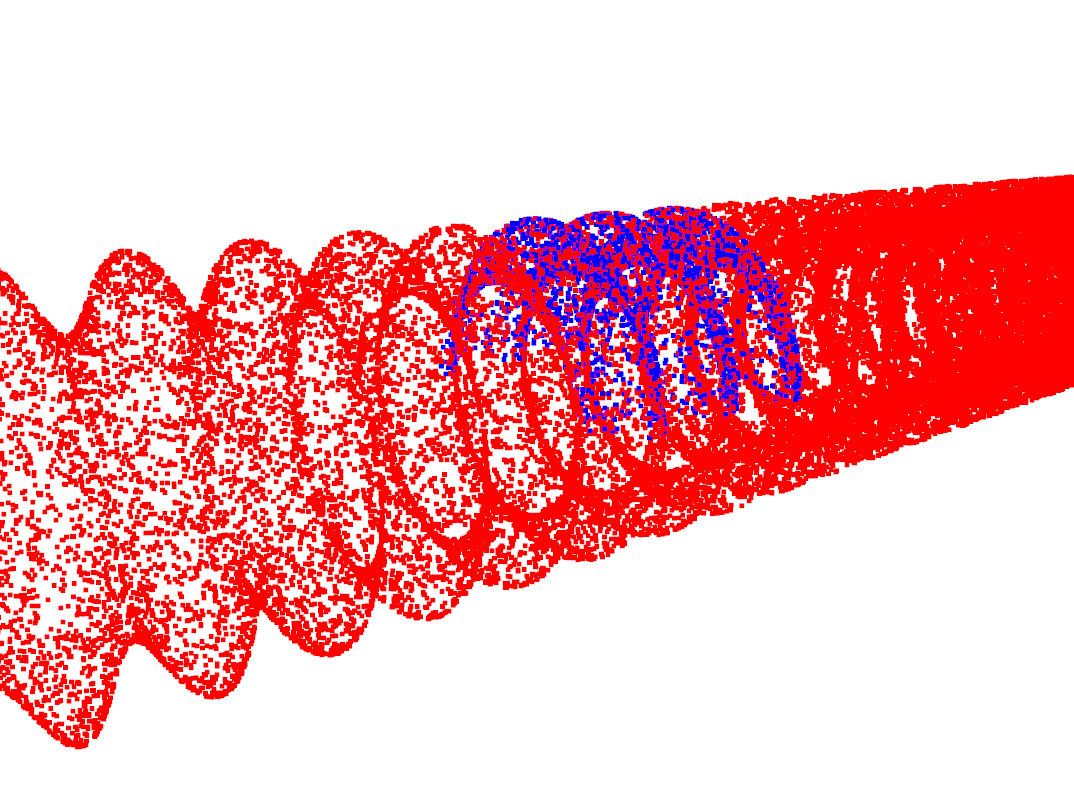}}
    \hfill
    \subfloat[Open Cylinder]{\includegraphics[height=2.5cm]{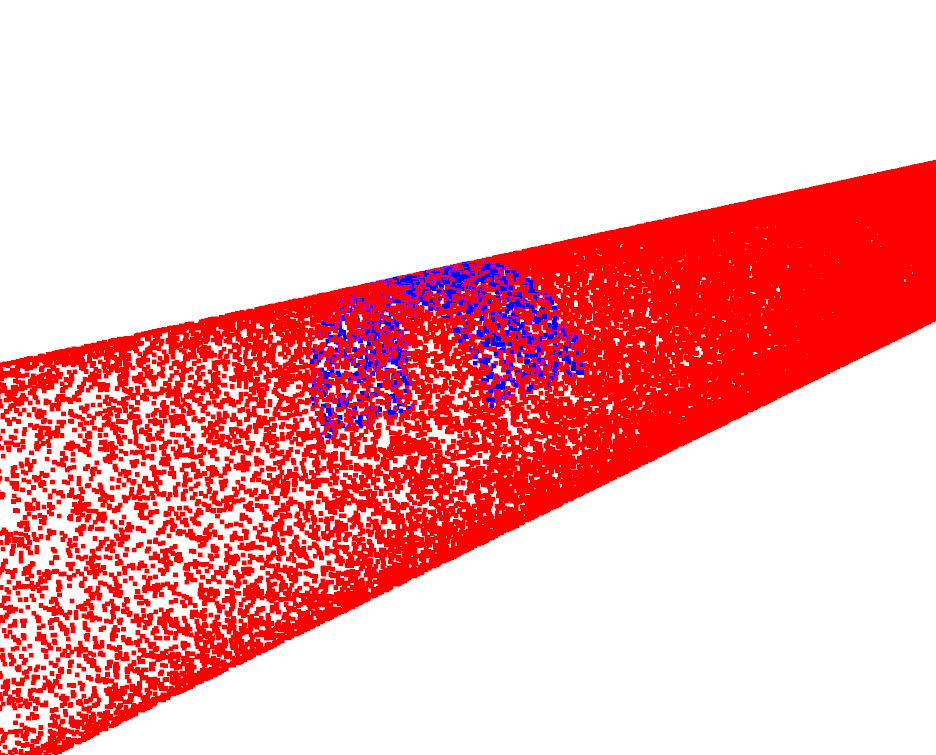}}
    
    \subfloat[Rotating Tunnel]{\includegraphics[height=2.5cm]{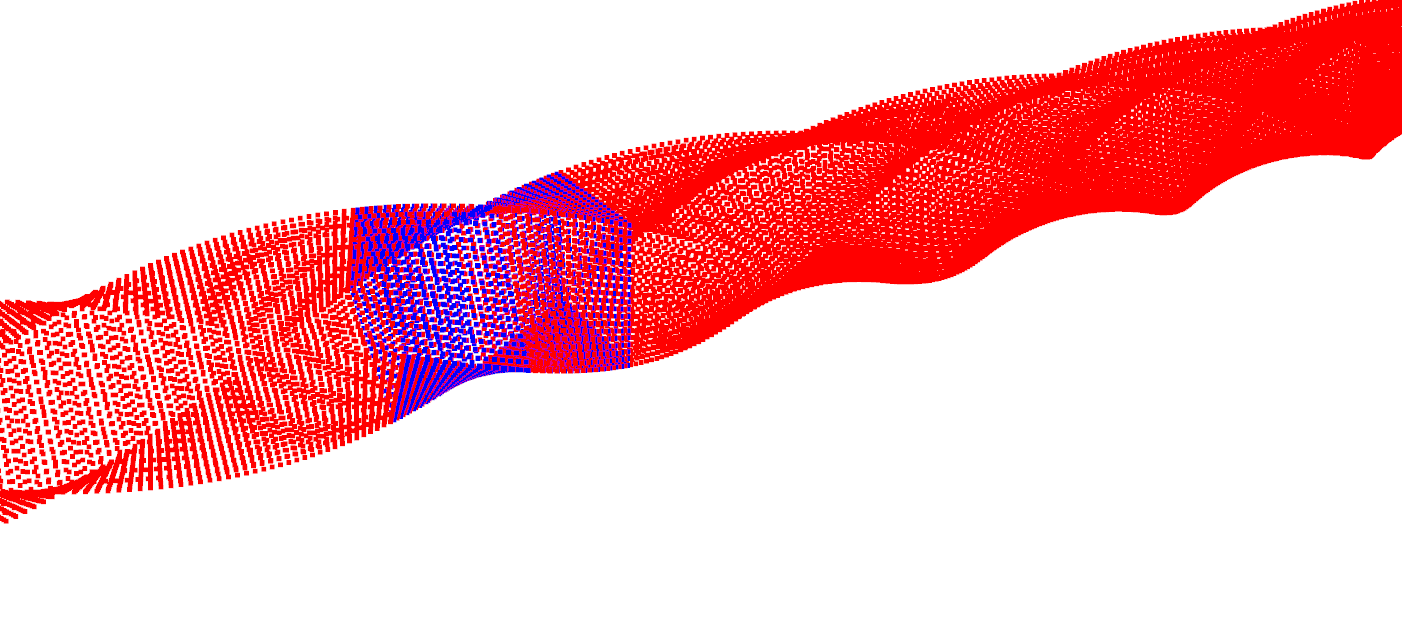}}
    
    \caption{The generated point clouds used for the synthetic data evaluation. Red: reference scan, blue: target scan.}
    \label{fig:generated_objects}
\end{figure}

The degeneracy of scan alignment is evaluated against the Hessian-based method on synthetic data generated from five 3D surface structures, each designed to exhibit a specific degeneracy scenario.
The scenarios are visualized in \cref{fig:generated_objects}.
The axes of the shapes are purposely misaligned with the $x-y-z$ axes to showcase the robustness of the approach.

We test each method's ability to identify the correct degenerate subspace, which is quantified using the proposed alignment quality metric \cref{eq:quality}.
The evaluation is performed for both ideal conditions and with Gaussian noise with distribution $\mathcal{N}(0, 0.02)$ added to each point of the point clouds along each axis.
The average distance between neighboring points in the generated point clouds is $0.05$ units.

\begin{table}[h!]\centering
   \caption{Alignment Error for Synthetic Datasets}\label{tab:alignment_results}
   \vspace{-1em}
%
\renewcommand{\arraystretch}{1.1}
\scalebox{1.1}{
\begin{tabular}{lrr}
   \toprule
   \multirow{2}{*}{\textbf{Scenario}} & \multicolumn{1}{c}{\textbf{Zhang's}} & \multicolumn{1}{c}{\textbf{LiLi}} \\
  & \multicolumn{1}{c}{\textit{Hessian-based}} & \multicolumn{1}{c}{\textit{Perturbation-based}} \\
   \midrule
   Plane & 0.033 (0.251) & 0.033 (\textbf{0.057}) \\
   Closed Cylinder & 0.030 (0.209) & \textbf{0.027} (\textbf{0.030}) \\
   Sinusoidal Cylinder & 0.066 (0.267) & \textbf{0.030} (\textbf{0.037}) \\   Open Cylinder & 0.382 (0.263) & \textbf{0.037} (\textbf{0.126}) \\
   Rotating Tunnel & 0.025 (0.069) & \textbf{0.017} (\textbf{0.046}) \\
   \bottomrule\noalign{\smallskip}
   \multicolumn{3}{p{0.8\linewidth}}{\scriptsize Values in the parentheses () denote results under noisy conditions.}
\end{tabular}
}

\end{table}
The alignment quality, measured by the median distance to the nearest neighbor of the extended point cloud to the reference point cloud, is summarized in \cref{tab:alignment_results}.
The results clearly demonstrate the robustness of the proposed method against noise and complex degeneracies.
For simpler structures like the Plane and Closed Cylinder, both compared methods perform well in noise-free conditions, but the proposed method consistently achieves lower error when the scans include noise.
The performance gap becomes significant in scenarios with more complex degeneracies, which involve coupled rotation and translation components, such as for the Open Cylinder.

\begin{figure}[h!]\centering
   \subfloat[Open Cylinder, no noise\label{fig:opened_perturb}] {
      \includegraphics[width=0.95\linewidth]{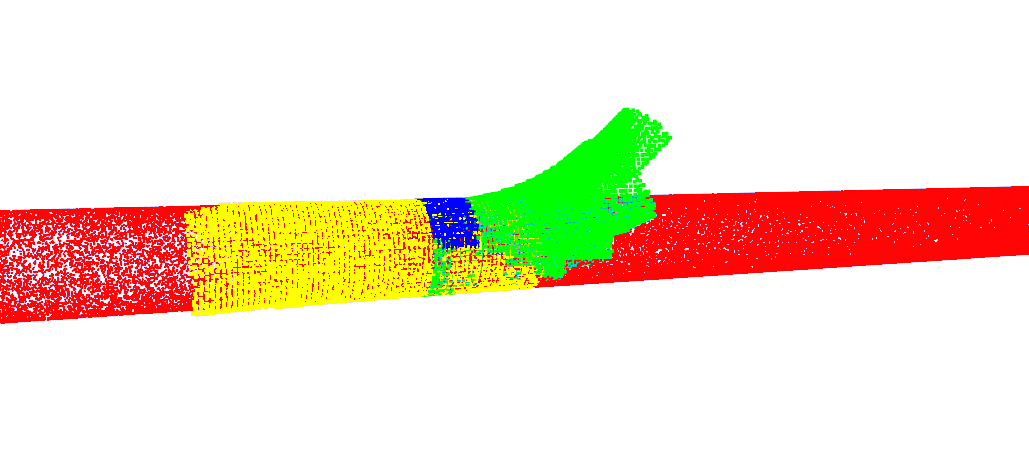}
   }

   \subfloat[Open Cylinder, with noise\label{fig:noise_opened_perturb}] {
      \includegraphics[width=0.95\linewidth]{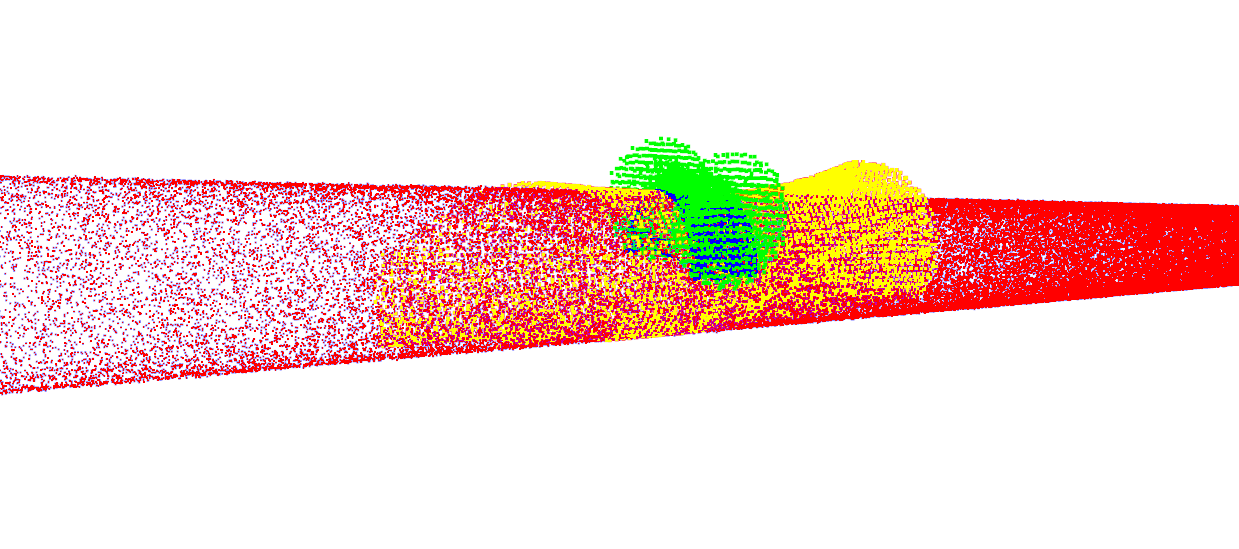}
   }
   \caption{
      The alignment of the Open Cylinder local scan (blue) to the reference scan (red). The extended scan, defined by~\cref{eq:extended_scan}, for the proposed method (yellow) aligns better than for the Hessian-based method (green).}
   \label{fig:opened_cylinder_alignment}
\end{figure}
We see an Open Cylinder scenario as the most challenging one, as the degeneracy subspace has a basis of $2$ degeneracies, formed from (1) translation along the cylinder axis and (2) rotation around the cylinder axis. For the local scan pose, rotation around the cylinder axis is a mixed translational-rotational motion in its local frame connected with the scan's center.
As shown in \cref{fig:opened_cylinder_alignment}, the proposed method correctly identifies these two coupled directions, allowing the local scan to slide along the surface while maintaining alignment with the reference scan in both noise-free and noisy conditions.
In contrast, the Hessian-based method performs worse in capturing the full local alignment degenerate subspace even in the absence of noise.
The error increases with noise, where the Hessian-based method completely misidentifies the degeneracy space, while our method remains robust due to its ability to account for data point re-association, inherent from the perturbation analysis.

\subsection{Effect of Detected Degeneracy on Localization Precision} \label{sec:results_localization}

The effect of the detected degeneracy on localization precision is demonstrated in a 3D LiDAR-Inertial localization system deployed in a tunnel environment, complemented by the wheel odometry.
When degeneracy is detected, the wheel odometry-derived local twist is projected into the degeneracy space.
It is complementary used in place of the LiDAR-based transformation in the degenerate subspace, similarly to the solution remapping procedure described in~\cite{zhang2016degeneracy}.

The employed LiDAR-Inertial odometry implementation is based on the LIO-SAM \cite{shan2020lio} system.
Notably, the standard LIO-SAM framework without degeneracy handling, enhanced by solution remapping, fails immediately upon entering the tunnel due to the lack of geometric constraints.
Therefore, we use two enhanced versions of the localization system: one using a Hessian-based \emph{Zhang's} method as a baseline, and one using the proposed Perturbation-based method \emph{LiLi}.

\begin{figure}[h!]\centering
   \includegraphics[trim=0 50 0 40, clip, width=0.7\columnwidth]{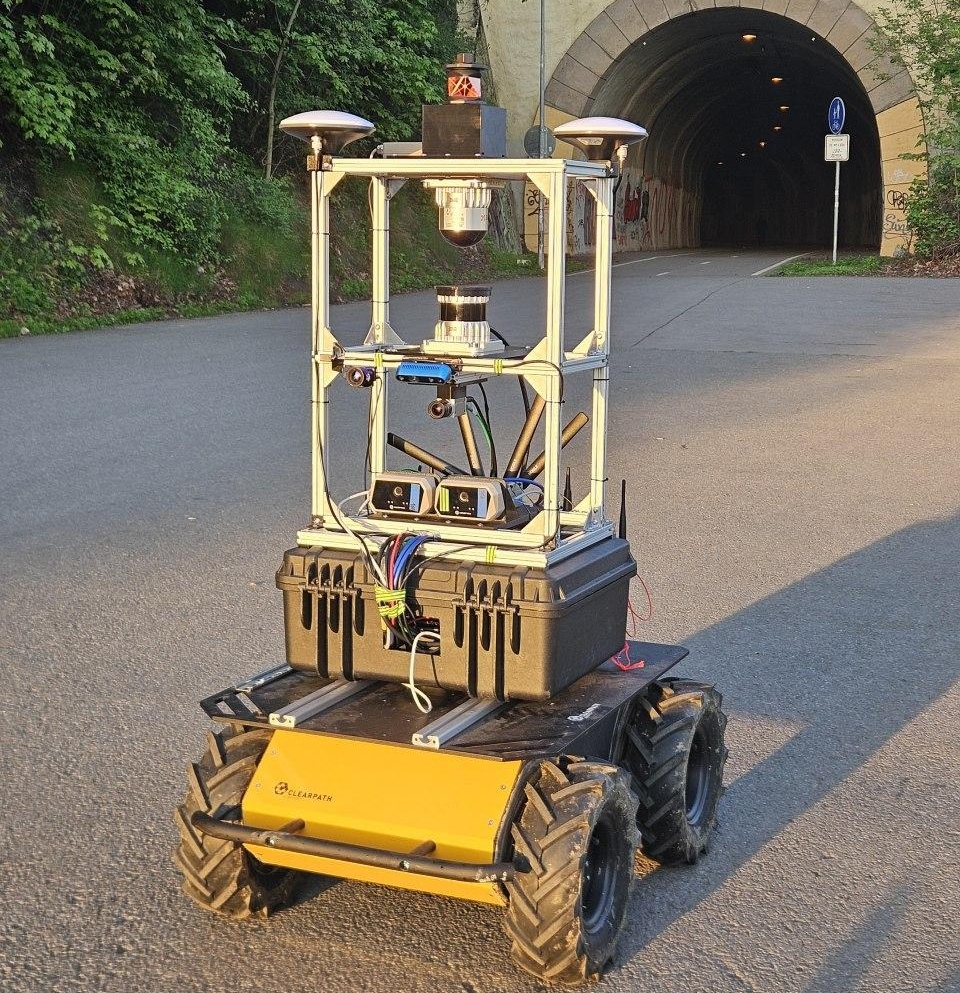}
   \caption{The Clearpath Husky A200 robot equipped with Ouster OS-1 3D LiDAR used for data collection at the tunnel entrance. The ground truth trajectories are captured by a total station that tracks the prism mounted on top of the sensory payload.}
   \label{fig:husky_robot_setup}
\end{figure}

\begin{figure}[h!]\centering
   \subfloat[Scan registration at entrance\label{fig:tunnel_entrance}] {
      \includegraphics[width=0.47\columnwidth]{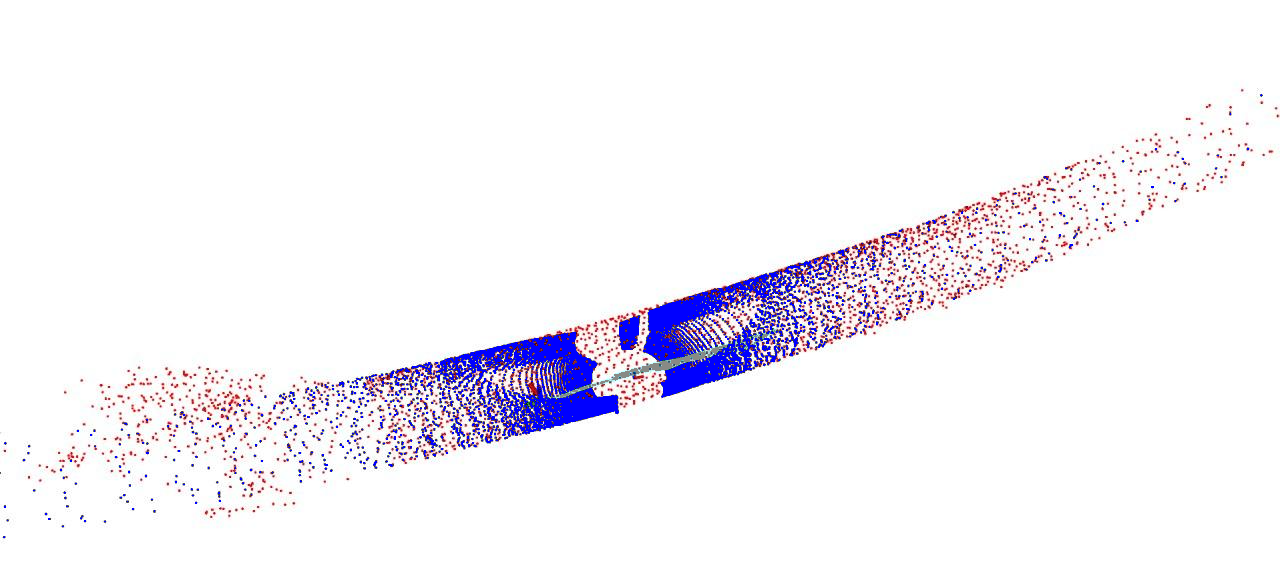}
   }
   \hfill
   \subfloat[Noisy map (Hessian-based)\label{fig:jacobian_map_noisy}] {
      \includegraphics[width=0.47\columnwidth]{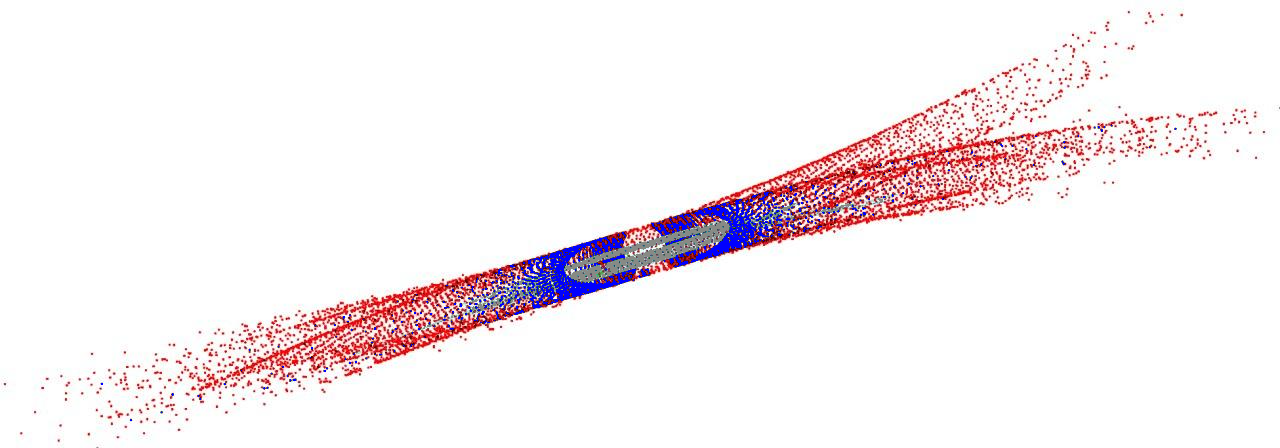}
   }
   \\[1ex]
   \subfloat[Full, clean map (Perturbation-based)\label{fig:proposed_map_clean}] {
      \includegraphics[angle=90, width=0.8\columnwidth]{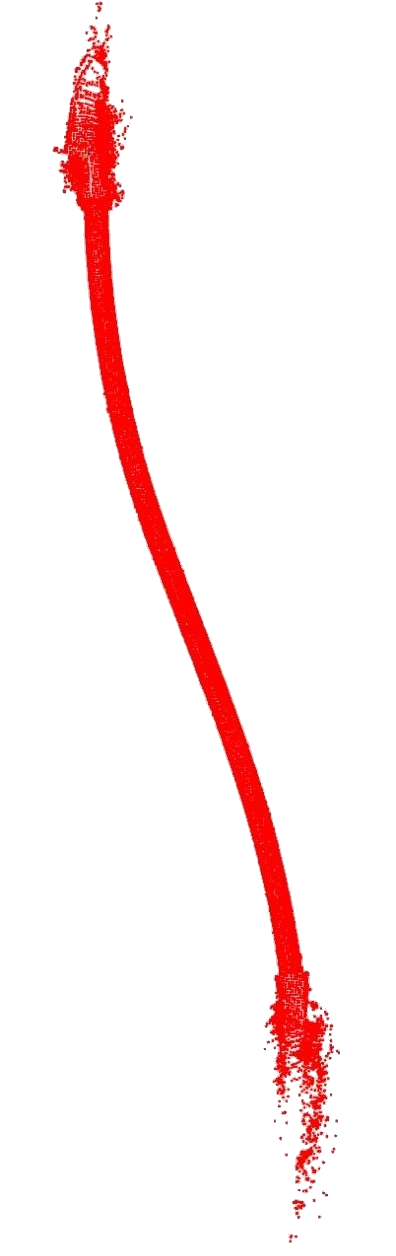} 
   }
   \caption{Map quality visualization.
   The accumulated map is shown in red, and the current LiDAR scan is blue.
   \protect\subref{fig:tunnel_entrance} A scan at the cluttered tunnel entrance.
   \protect\subref{fig:jacobian_map_noisy} The map, resulting from the failed localization of Hessian-based Zhang's method.
   \protect\subref{fig:proposed_map_clean} The map, resulting from the proposed LiLi method.}
   \label{fig:map_comparison}
\end{figure}

\begin{table}[h!]\centering
   \caption{Localization Results in the Tunnel Environment}\label{tbl:localization_results}
   \vspace{-1em}
%
\renewcommand{\arraystretch}{1.1} 
\scalebox{1.1}{
   \begin{tabular}{clrr}
      \toprule
      \textbf{Trial} & \textbf{Method} & \textbf{ATE} [m] & \textbf{RPE} [m] \\
      \midrule
      \multirow{2}{*}{Trial 1} & Zhang's & 0.370 & 0.820 \\
      & LiLi & \textbf{0.290} & \textbf{0.470} \\
      \noalign{\smallskip}
      \multirow{2}{*}{Trial 2} & Zhang's & Failed & Failed \\
      & LiLi & \textbf{0.536} & \textbf{1.020} \\
      \bottomrule
   \end{tabular}
}

\end{table}
We showcase the results on a dataset we collected in a~\SI{260}{\meter} long curved tunnel, a structurally degenerate environment, 3D map of which can be observed in~\cref{fig:proposed_map_clean}.
The experiment was conducted using a Clearpath Husky A200 robot, shown in~\cref{fig:husky_robot_setup}, equipped with an Ouster OS-1 \num{128}-beam 3D LiDAR featuring a built-in \num{6}-axis IMU.

The 3D ground truth position of the robot is captured by a total station Leica TS16, which allows for computing \emph{Absolute Trajectory Error} (ATE)~\cite{wulf_atebenchmarking_jfr08} and the translational part of the \emph{Relative Pose Error} (RPE)~\cite{kummerle2009_rpeOrigin} as performance indicators of the localization system. We compute the RPE as a root-mean-square error (RMSE) over all relative position errors for all the trajectory intervals with a path length of \SI{100}{\meter}.

Two consecutive trials are considered for the evaluation: a~\SI{260}{\meter} long single pass through the tunnel, and a \SI{430}{\meter} long round trip in the tunnel, further referred to as Trial~1 and Trial~2, respectively.
The performance results are summarized in \cref{tbl:localization_results}.

\begin{figure}[htb]\centering
   \includegraphics[width=\columnwidth]{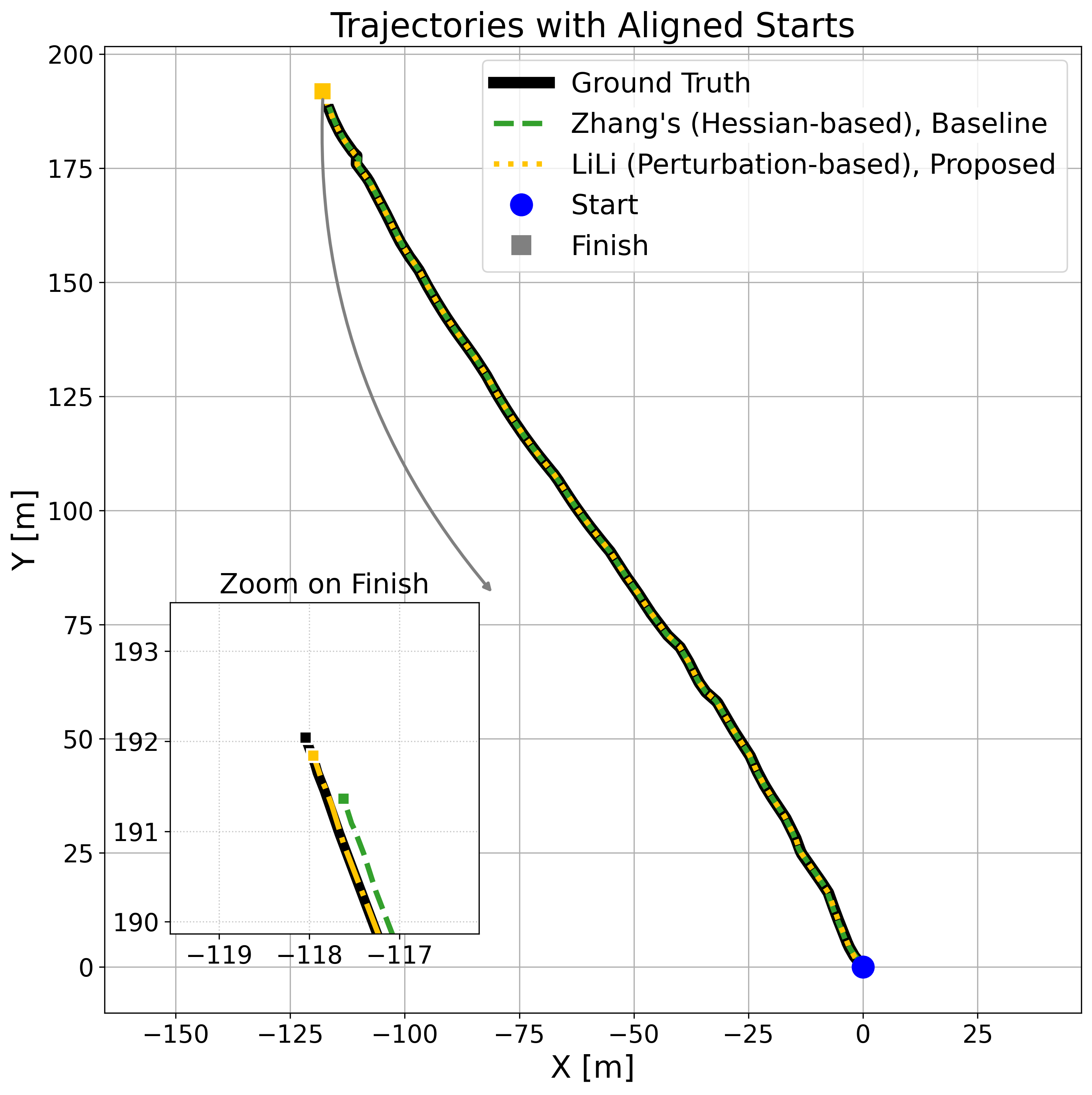}
   \caption{\label{fig:aligned_trajectories_plot}
   Aligned trajectories for Trial 1, the proposed method, the Hessian-based baseline, and the ground truth.
   The inset provides a zoomed-in view of the final positions, showing the superior accuracy of the proposed method.
   }
\end{figure}
In a single pass Trial 1, the proposed method achieves the ATE of \SI{0.290}{\meter} and the RPE of \SI{0.470}{\meter} per \SI{100}{\meter}, outperforming the baseline, as shown in \cref{fig:aligned_trajectories_plot}.

The performance difference of the proposed and baseline methods is pronounced in round-trip Trial 2.
The Hessian-based method failed to maintain localization at the cluttered tunnel entrance, see \cref{fig:map_comparison}, leading to a corrupted map and complete failure as a result.
In contrast, the proposed method successfully detected the degeneracy and completed the entire trajectory with a~high accuracy.


\section{Discussion} \label{sec:discussion}
The evaluation results on the synthetic data highlight a~key strength of the proposed method: its robustness in noisy conditions and in scenarios with complex, mixed degeneracies. 
While the Hessian-based method performs competitively in simple, noise-free cases, its performance degrades significantly with the introduction of noise or geometric complexity.
We attribute our method's superior performance to its ability to account for data point re-association through its perturbation-based analysis.

The robustness translates directly to the real-world localization trials.
The cluttered tunnel entrance, combined with the tunnel's structural degeneracy, represents a challenging scenario where the Hessian-based method failed utterly in the longer Trial 2.
Its failure to detect the degeneracy at the entrance led to map corruption, and the corrupted map prevented further successful detection of degeneracy as it was a cluttered environment. 
In contrast, the proposed method successfully handles challenges of the experimental tunnel environment, maintaining accurate localization throughout both trials.
Moreover, in case of the Trial 1, where both methods complete the trajectory, the proposed method achieves a \SI{21}{\percent} lower ATE and a \SI{43}{\percent} lower RPE than the baseline, showing that the proposed method is not only more robust, but also more accurate in estimating degeneracy directions in the real-world scenarios. 
It demonstrates that the resilience to noise and complex degeneracies observed in the synthetic tests is critical for reliable performance in real-world applications.

Despite its strong performance, the proposed method has a few limitations. 
We run the localization system at \SI{10}{\hertz}, which is sufficient for many robotic applications, but may not be suitable for high-speed scenarios.
Additionally, it may incorrectly identify degeneracies in highly repetitive environments and is primarily designed for surface-like point clouds rather than dense, volumetric data.
A future direction is to mitigate the above limitations, utilize the proposed degeneracy description method for global map optimization, and explore its utility in SLAM applications.

\section{Conclusion} \label{sec:conclusion}

In the presented work, we address challenges of detecting and characterizing degeneracies in LiDAR scan alignment.
In particular, in structurally degenerate and noisy scenarios, the scan can move along specific combined translational and rotational directions without affecting alignment metrics.
We proposed a novel degeneracy detection algorithm called LiLi that leverages perturbations of the optimized pose to describe the full subspace of degenerate transformations $SE(3)$ using the basis directions from the Lie algebra $\mathfrak{se}(3)$.

The presented evaluation on synthetic data demonstrates that the proposed method significantly outperforms the state-of-the-art Hessian-based approach.
Specifically, under noisy conditions and in geometrically complex scenarios, the alignment is improved by over \SI{50}{\percent}. 
The practical value of the method's robustness is further confirmed through experimental validation based on integration of the method into a LiDAR-Inertial odometry system.
In the real-world, structurally degenerate tunnel environment, the proposed approach enabled continuous and accurate localization over a \SI{430}{\meter} long round-trip trajectory, a task where the baseline system equipped with a Hessian-based detector failed, and outperformed the baseline accuracy in cases where both methods succeeded.
The achieved result underscores the method's ability to prevent catastrophic localization failure and ensures the theoretical basis for reliable navigation in real-world applications.

\section{Acknowledgments} \label{sec:acknowledgments}
The authors gratefully acknowledge the assistance of Large Language Models, specifically GitHub Copilot and Gemini, in generating code for visualization and integration into the LiDAR-Inertial odometry system.


\bibliographystyle{IEEEtranDOI}
\bibliography{main}

\end{document}